\def\year{2019}\relax
\documentclass[letterpaper]{article} 
\usepackage{aaai19}  
\usepackage{times}  
\usepackage{helvet}  
\usepackage{courier}  
\usepackage{url}  
\usepackage{graphicx}  
\usepackage{booktabs}
\usepackage{multirow}
\usepackage{amssymb}
\usepackage{amsmath}
\usepackage{color}

\DeclareMathOperator*{\argmaxA}{arg\,max}

\begin{document}
%
\title{Unseen Word Representation\\by Aligning Heterogeneous Lexical Semantic Spaces}

\author{
Victor Prokhorov\textsuperscript{1},
Mohammad Taher Pilehvar\textsuperscript{1,2},
Dimitri Kartsaklis\textsuperscript{3}\thanks{Contributed to this work when he was with DTAL, Cambridge.},
Pietro Li\`o\textsuperscript{4},
Nigel Collier\textsuperscript{1} 
\vspace{0.1cm}
\\
\textsuperscript{1}Department of Theoretical and Applied Linguistics, University of Cambridge\\
\textsuperscript{2}School of Computer Engineering, Iran University of Science and Technology, Tehran, Iran\\
\textsuperscript{3}Apple, Cambridge, UK\\
\textsuperscript{4}Department of Computer Science, University of Cambridge
\vspace{0.1cm}
\\
vp361@cam.ac.uk, pilehvar@iust.ac.ir, dkartsaklis@apple.com, pl219@cam.ac.uk, nhc30@cam.ac.uk
}


\newcommand\BibTeX{B{\sc ib}\TeX}
\newcommand{\newcite}[1]{\citeauthor{#1} (\citeyear{#1})}
\newcommand{\citep}[1]{\citeauthor{#1} (\citeyear{#1})}

\maketitle
\begin{abstract}
Word embedding techniques heavily rely on the abundance of training data for individual words.
Given the Zipfian distribution of words in natural language texts, a large number of words do not usually appear frequently or at all in the training data.
In this paper we put forward a technique that exploits the knowledge encoded in lexical resources, such as WordNet, to induce embeddings for unseen words.
Our approach adapts graph embedding and cross-lingual vector space transformation techniques in order to merge lexical knowledge encoded in ontologies with that derived from corpus statistics.
We show that the approach can provide consistent performance improvements across multiple evaluation benchmarks: \textit{in-vitro}, on multiple rare word similarity datasets, and \textit{in-vivo}, in two downstream text classification tasks.
\end{abstract}

\section{Introduction}



Word embeddings can be seamlessly integrated into various NLP systems, effectively enhancing their generalisation power \cite{survey:2018}.
However, the distributional approach to the semantic representation of words, either in its conventional count-based form or the recent neural-based paradigm, relies on a multitude of occurrences for each individual word to enable accurate representations.
As a result, these corpus-based methods are unable to provide reliable representations for words that are infrequent or unseen during training, such as domain-specific terms.
This is the case even if massive corpora are used for training, such as the Wikipedia corpus.\footnote{\label{foot:vocab}In the 2015 Wikipedia dump corpus with around 1.6B tokens, there are slightly over 1.9M word types with at least three occurrences. Of these word types, more than 80\% appear at most 50 times in total, whereas more than two thirds of words in the vocabulary have frequency $\le$ 20.}

To address the unseen word representation problem, several techniques have been proposed.
Earlier works have mainly focused on morphologically complex words \cite{luong-socher-manning:2013,Botha2014,soricut-och:2015}, whereas 
more recently, character-based and subword unit information has garnered a lot of attention \cite{fasttext-subword:TACL999}.
Despite their success, these models make two assumptions around the unseen word: (1) variations of the word exist in the training corpus (for instance, occurrences of 
\textit{track}---or even \textit{untrack}---should exist to induce embeddings for \textit{untracked}); and (2) the semantics of the word can be estimated based on its subword units (which might not hold for single-morpheme words, e.g., \textit{galaxy}, or for exocentric compounds, e.g., \textit{honeymoon}). As a result, they fall short of effectively representing the semantics of unseen single-morpheme words for which no variation has been observed during training, essentially ignoring most of the rare domain-specific entities which are crucial for NLP systems when applied to those domains.


\begin{figure*}[t!]
\begin{center}
	\includegraphics[trim = 0mm 10mm 0mm 0mm,,scale=0.21]{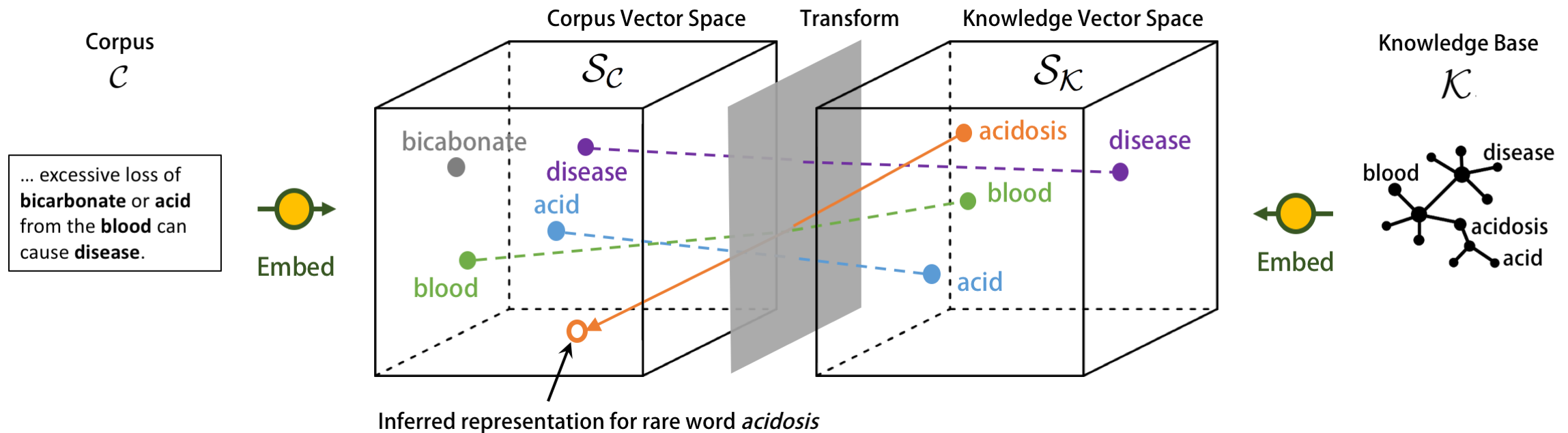}
\end{center}
 \caption{Our coverage enhancement procedure. The dashed lines represent semantic bridges and the solid line represents a rare word that is projected from the knowledge vector space to the corpus vector space.}
 \label{fig:procedure}
\end{figure*}

Furthermore, distributional techniques generally ignore the lexical knowledge encoded in dictionaries, ontologies, or other lexical resources.
There exist hundreds of high coverage and domain-specific lexical resources which contain valuable information for infrequent words.
Recently, various embedding induction techniques have attempted to leverage lexical resources, such as WordNet \cite{pilehvar-collier:2017:EACLshort,embeddings-on-fly:2017} or Wikipedia \cite{Angeliki2017-LAZMWM}.
Despite their success, they either rely on word definitions (glosses) or related words extracted from the the lexical resource while ignoring the knowledge encoded in the semantic structure.
Here, we present a methodology that exploits the semantic structure of the lexical resource for unseen word representation.
The technique first embeds a knowledge base into a vector space and then maps the embedded words from this space to a corpus-based space, in order to expand the vocabulary of the latter with additional representations for rare and unseen words.
To our knowledge, it is the first time that vector space transformation techniques, which are widely used in multilingual settings, are leveraged for aligning heterogeneous monolingual spaces.
We evaluate the reliability of our approach on several datasets across multiple tasks: six datasets for word similarity measurement and eight sentiment analysis and topic categorization datasets.
Experimental results show that, unless ample occurrences exist in the training data, we can compute more reliable embeddings than the ones generated by state-of-the-art corpus based embedding techniques.



\section{Methodology}
\label{sec:method}


Figure \ref{fig:procedure} illustrates our procedure for enriching an existing corpus vector space $\mathcal{S}_\mathcal{C}$ based on the lexical knowledge in an external knowledge base (KB) $\mathcal{K}$.
The proposed algorithm mainly relies on techniques from two research areas: graph embedding and vector space transformation.
Two main steps are involved in the process.
First, it views $\mathcal{K}$ as a knowledge graph and transforms it to a vector space representation ($\mathcal{S}_\mathcal{K}$) by leveraging graph embedding techniques (Section \ref{sec:kb_embedding}).
Then, it aligns the two vector spaces, i.e., KB- ($\mathcal{S}_\mathcal{K}$) and corpus-based ($\mathcal{S}_\mathcal{C}$), by using vector space transformation algorithms (Section \ref{sec:transformation}).
As a result of this alignment, new embeddings are induced for unseen words in $\mathcal{S}_\mathcal{C}$.
In our toy example in Figure 1, the term \textit{acidosis} is missing from the vocabulary of $\mathcal{S}_\mathcal{C}$ but it is covered by the knowledge base $\mathcal{K}$.
First, a graph embedding algorithm is used to embed $\mathcal{K}$, represented as a graph, into a vector space $\mathcal{S}_\mathcal{K}$.
Then, based on common clues from the two spaces, a transformation function is learnt in order to map the vectors across the two spaces.
The transformation function (from the embedded KB space $\mathcal{S}_\mathcal{K}$ to the corpus space $\mathcal{S}_\mathcal{C}$) allows us to project the vector for \textit{acidosis} to the latter space, hence inducing a new representation for the word.

\subsection{Prerequisites}
\label{sec:prereq}

The knowledge base  embedding algorithms used in our procedure require $\mathcal{K}$ to be representable as a semantic network (knowledge graph).
In our experiments, we used WordNet 3.0 \cite{Fellbaum:98} as external knowledge base.
The resource contains around 120K groups of synonyms, referred to as {\em synsets}, which are connected to each other by means of around 200K lexical semantic relations, such as hypernymy and meronymy. 
Thanks to these relations, WordNet can be readily viewed as a semantic network.
We further enrich the network by connecting a synset to all other synsets that appear in its disambiguated gloss\footnote{\url{wordnet.princeton.edu/glosstag.shtml}}.
This approach more than doubles the number of edges in WordNet's semantic network.
As for the corpus vector space, any distributional semantic representation can be used. 
In our experiments, we opted mainly for word embeddings (rather than conventional count-based representations) due to their popularity.

Our procedure requires two additional conditions.
Let $V_\mathcal{K}$ and $V_\mathcal{C}$ be the respective vocabularies of knowledge base and corpus vector spaces.
The first condition to be met is that $V_\mathcal{K}$ and $V_\mathcal{C}$ should have overlapping words, i.e., $V_\mathcal{K} \cap V_\mathcal{C} \ne \phi$.
This is required for enabling the alignment of the two spaces (to be discussed in Section \ref{sec:transformation}).
The second condition is that the knowledge base $\mathcal{K}$ has to provide lexical knowledge for unseen or infrequent words in the corpus vector space.
Thanks to the abundance of knowledge bases
and the long tail of words in distributional representations, this condition is not difficult to be fulfilled.

\subsection{Knowledge Base Embedding}
\label{sec:kb_embedding}

The proposed coverage enhancement procedure starts by transforming the lexical knowledge representation in the knowledge base $\mathcal{K}$ to a form which is comparable to the corpus-based representation $\mathcal{S}_\mathcal{C}$.
To this end, we \textit{embed} the structural lexico-semantic knowledge of $\mathcal{K}$ into a vector space $\mathcal{S}_\mathcal{K}$.

We opted for node2vec\footnote{\url{https://github.com/snap-stanford/snap/tree/master/examples/node2vec}} \cite{node2vec:2016}, a random walk based graph embedding technique which has proven its potential in the reliable representation of graph nodes.
Given a graph $G$, the algorithm first generates a stream of artificial ``sentences'' by performing a series of random walks over $G$.
Each such ``sentence'' contains a sequence of ``words" (i.e, vertices) such that consecutive words correspond to neighbouring vertices in $G$.
Analogously to the natural language text in which semantically similar words are expected to appear in similar contexts, an artificial sentence encodes local information for a node from the graph by placing topologically close vertices in similar contexts.
Representations are then computed for individual vertices by taking a similar objective to the Skip-gram model \cite{Mikolovetal:2013}, i.e., by maximizing
$\prod_{j=i-z, j \ne i}^{i+z} \mathbf{Pr}(w_j|w_i)$ which is the probability of a word $w_i$ given its context, where $z$ is the window size or the length of the random walk.
The only difference from the original Skip-gram model lies in the way input ``sentences'' are constructed.


In our experiments, we set the parameters of node2vec as follows: walk length to 100, window size to 10, and embedding dimensionality to 100. 
To decide on these parameters, we carried out experiments on the MTURK-771 dataset \cite{mturk-771}.
Also, note that nodes in the semantic graph of WordNet represent synsets. Hence, a polysemous word would correspond to multiple nodes.
In our word similarity experiments (Sections \ref{sec:rw_exp} and \ref{sec:sim_exp}) we use the \textit{MaxSim} assumption of \newcite{ReisingerMooney:2010} in order to map words to synsets: the similarity of two words is computed as that of their closest associated meanings.
In the downstream experiment (Section \ref{sec:downstream}), we compute a single word vector as the average of its corresponding synsets' vectors.

\subsection{Vector Space Alignment}
\label{sec:transformation}

Once the lexical resource $\mathcal{K}$ is represented as a vector space $\mathcal{S}_\mathcal{K}$, we project it to $\mathcal{S}_\mathcal{C}$ in order to improve the word coverage of this space with additional words from $\mathcal{S}_\mathcal{K}$.
In this procedure we make two assumptions. Firstly, the two spaces provide reliable models of word semantics; hence, the relative within-space distances between words in the two spaces are comparable.
Secondly, there exists a set of shared words between the two spaces (also mentioned in Section \ref{sec:prereq});
we refer to these words as \textit{semantic bridges}.


For this transformation we opted for Canonical Correlation Analysis \cite[ CCA]{faruqui-dyer:2014,upadhyay-EtAl:2016}, which is widely used for the projection of spaces belonging to different languages with the purpose of learning multilingual semantic spaces.\footnote{We also performed experiments with least squares regression \cite{Mikolovetal:13,Dinuetal:2014}. However, CCA proved to be consistently better. We do not report LS results due to lack of space.}
The model receives as input two vector spaces for two different languages and a seed lexicon for that language pair, and learns a linear mapping between the two spaces.
Ideally, words that are semantically similar across the two languages will be placed in close proximity to each other in the projected space.

Specifically, let $\mathcal{S}'_\mathcal{C} \subset \mathcal{S}_\mathcal{C}$ and $\mathcal{S}'_\mathcal{K} \subset \mathcal{S}_\mathcal{K}$ be the corresponding subsets of semantic bridges, i.e., words that are monosemous according to the WordNet sense inventory, for corpus and KB spaces, respectively.
Note that $\mathcal{S}'_\mathcal{C}$ and $\mathcal{S}'_\mathcal{K}$ form matrices that contain representations for the same set of words, i.e., $|\mathcal{S}'_\mathcal{C}| = |\mathcal{S}'_\mathcal{K}|$.
%
%
CCA finds a linear combination of dimensions in $\mathcal{S}_\mathcal{C}$ and $\mathcal{S}_\mathcal{K}$ which have maximum correlation with each other.
Given two column vectors $\mathcal{S}'_\mathcal{C}$ and $\mathcal{S}'_\mathcal{K}$ of embeddings in the two spaces, CCA computes vectors $w_\mathcal{C}$ and $w_\mathcal{K}$ such that the random variables $w_\mathcal{C} \mathcal{S}'_\mathcal{C}$ and $w_\mathcal{K}\mathcal{S}'_\mathcal{K}$ maximize the correlation $\rho (w_\mathcal{C} \mathcal{S}'_\mathcal{C}, w_\mathcal{K} \mathcal{S}'_\mathcal{K})$:

\begin{equation*}
\begin{split}
    w^*_\mathcal{C}, w^*_\mathcal{K} &= \mathbf{CCA}(\mathcal{S}'_\mathcal{K},\mathcal{S}'_\mathcal{C})\\
                              &= \argmaxA_{w_\mathcal{C},w_\mathcal{K}} \rho (w_\mathcal{C} \mathcal{S}'_\mathcal{C}, w_\mathcal{K} \mathcal{S}'_\mathcal{K})\\
                        &=
            \argmaxA_{w_\mathcal{C},w_\mathcal{K}} \frac{w_\mathcal{C} \Sigma_{\mathcal{C}\mathcal{K}} w_\mathcal{K}}{\sqrt{w_\mathcal{C} \Sigma_\mathcal{C} w_\mathcal{C}^T} \sqrt{w_\mathcal{K} \Sigma_\mathcal{K} w_\mathcal{K}^T}} 
\end{split}
\end{equation*}

\noindent
where 
$\Sigma_X$ and $\Sigma_{X,Y}$ denote covariance and cross-covariance, respectively.
Note that the maximization is invariant to scaling of $w_\mathcal{C}$ and $w_\mathcal{K}$. Hence, we can have a constraint for unit variance:

~
\vspace{-0.3cm}
\begin{equation*}
w^*_\mathcal{C}, w^*_\mathcal{K} = \argmaxA_{w_\mathcal{C} \Sigma_\mathcal{C} w_\mathcal{C}^T=w_\mathcal{K} \Sigma_\mathcal{K} w_\mathcal{K}^T=1} w_\mathcal{C} \Sigma_{\mathcal{C}\mathcal{K}} w_\mathcal{K}
\end{equation*}

The dimensionality of the resultant space in our experiments is $\min(d_\mathcal{C},d_\mathcal{K})=d_\mathcal{K}=100$, where $d_\mathcal{C}$ and $d_\mathcal{K}$ are the dimensionalities of the corpus and KB spaces, respectively. 
An additional constraint forces these projections to be uncorrelated.
The enhanced space $\mathcal{S}^*$ is obtained as the union of $w_\mathcal{C} \mathcal{S}_\mathcal{C}$ and $w_\mathcal{K} \mathcal{S}_\mathcal{K}$.
Note that this procedure is slightly different from the one illustrated in Figure \ref{fig:procedure}. 
The enriched space is a third space which is independent from the two initial spaces $\mathcal{S}_\mathcal{K}$ and $\mathcal{S}_\mathcal{C}$.

As for the seed lexicon (the set of semantic bridges $\mathcal{S}'_\mathcal{C}$ and $\mathcal{S}'_\mathcal{K}$), we used the set of monosemous words in the WordNet's vocabulary which are deemed to have the most reliable semantic representations in the corpus vector space.
Of the 155K words in WordNet's vocabulary, around 128K are monosemous, which provides us with a large set of semantic bridges to use for the alignment step.
However, in our experiments we found that a small subset of 5K semantic bridges is enough for achieving reliable transformations.

\paragraph{Graph embedding and space alignment.} 
For this work we experimented with node2vec. 
We note that there is a rich literature for graph embeddings \cite{Caietal:2017}.
A series of algorithms first construct an adjacency matrix of the graph and obtain embeddings by directly factorising this matrix \cite{Caoetal:2015,Roweis2000}, whereas others 
employ deep learning techniques, such as autoencoders \cite{Wang:2016}.
Relation embedding techniques such as TransE \cite{Bordesetal:2013} and HOLE \cite{Nickeletal:2016} are not suitable candidates for our purpose since their focus is rather on the embedding of edges (as opposed to nodes).
As noted before, for the space alignment we experimented with CCA which is a linear model of projection.
We leave the evaluation of non-linear transformation techniques, such as Kernel CCA \cite{Akaho:2006} and Deep CCA \cite{galen2013-deep-cca}, and other graph embedding techniques, to future work.

\section{Experiments}


In this section we provide three different sets of experiments that were carried out to evaluate the reliability of our rare word embedding induction technique (which we will refer to as \textbf{\textsc{Align}}).
First, we report results for in-vitro evaluations on the Stanford Rare Word similarity dataset (Section \ref{sec:rw_exp}) and in a simulated rare word similarity setting (Section \ref{sec:sim_exp}).
We then verify the reliability of our induced embeddings in two downstream NLP tasks, sentiment analysis and topic categorization.
This experiment is detailed in Section \ref{sec:downstream}. The code used in our experiments will be released to allow future experimentation and comparison.\footnote{\url{https://github.com/VictorProkhorov/AAAI2019}}

\subsection{Rare Word Similarity}
\label{sec:rw_exp}

The Stanford Rare Word (RW) Similarity dataset \cite{luong-socher-manning:2013} has been regarded as a standard benchmark for evaluating embedding induction techniques.
The dataset comprises 2034 pairs of infrequent words, such as \textit{ulcerate}-\textit{change} and \textit{nurturance}-\textit{care}. 
In the first evaluation, we use this benchmark to compare our model against recent rare word representation techniques.

\paragraph{Experimental setup.} 
We experimented with two sets of word2vec \cite{Mikolovetal:2013} embeddings trained on two different corpora: (1) 
\textsc{w2v-gn}, the Google News (vocab: 3M, dim: 300)\footnote{\url{code.google.com/archive/p/word2vec/}},
and (2) \textsc{w2v-wp}, the Wikipedia corpus \cite{WikiCorp} (vocab: 2.4M, dim: 300).
As for comparison systems, we benchmark our results against four other approaches: (1) \textbf{SemLand} \cite{pilehvar-collier:2017:EACLshort} which extracts for an unseen word the set of its semantically related words from WordNet and induces an embedding by combining their embeddings; (2) the \textbf{Additive} model of \newcite{Angeliki2017-LAZMWM} which takes the unseen word's definition as semantic clue and induces an embedding by adding (averaging) the embeddings of content words in the defintion; (3) \textbf{LSTM}-based strategy of \newcite{embeddings-on-fly:2017} which is a more complex version of the additive model that relies on an LSTM network which receives as its input the WordNet definition of the unseen word; and (4) \textbf{FastText} \cite{fasttext-subword:TACL999} which computes a word embedding by combining the embeddings of its sub-word character n-grams (see Section \ref{sec:related_work} for more details).

\begin{table}[t!]
\setlength{\tabcolsep}{8pt}
\begin{center}
\scalebox{1}
{
\begin{tabular}{l cc c cc}
\toprule

\bf \multirow{2}{*}{Embedding} & \multicolumn{2}{c}{{\bf \textsc{w2v-gn}}} & & \multicolumn{2}{c}{{\bf \textsc{w2v-wp}}} \\
\cmidrule(lr){2-3}
\cmidrule(lr){5-6}

 &  $r$ &   $\rho$ & &   $r$ &   $\rho$ \\

\midrule
{\bf \textsc{w2v-gn}}   
&	{0.44}	&0.45   && 0.41 &	0.43 \\

+ Additive & 0.46	& 0.48	&& 0.41 &	0.43 \\

+ SemLand & 0.48	& 0.51	&& 0.39 &	0.40 \\

+ LSTM &0.48  & 0.50    &&0.40 &0.40 	\\

+ \textsc{Align}   & 0.48	& 0.48	&&0.42 &0.42	\\






\bottomrule

\end{tabular}
}
\end{center}
\caption{\label{table:rw-sim}Pearson ($r$) and Spearman ($\rho$) correlation for our approach (\textsc{Align}) on the RW dataset with two pre-trained sets of word embeddings, before and after enhancement with various methods. 
FastText\textsc{-wp} (trained on the Wikipedia corpus): $r=0.44$, $\rho=0.44$ and node2vec (without any alignment and independent from corpus embeddings): $r=0.16$, $\rho=0.16$.
}
\end{table}

\paragraph{Results.} 
 Table \ref{table:rw-sim} shows correlation performance on the dataset for the two pre-trained word embeddings, in their initial form and when enhanced with additional induced word embeddings.
Among the two initial embeddings, \textsc{w2v-gn} provides a lower coverage (173 out-of-vocabulary words vs. 88 for \textsc{w2v-wp}) despite its larger vocabulary (3M vs. 2.4M).
All enhanced embeddings attain near full coverage (over 99\%), thanks to the vocabulary expansion offered by WordNet.
Our approach (\textsc{Align}) produces competitive performance across the two settings and according to both Pearson and Spearman correlation metrics. 
The performance ($r=0.16$, $\rho=0.16$) of node2vec, when independently applied to this dataset, is notably lower than that of the initial corpus embeddings. 
However, it is interesting to note that these non-optimal embeddings can better the performance of corpus embeddings when combined with them, showing
the complementarity of the two sources of information.

\paragraph{Comparison with FastText.}
FastText proves competitive on the dataset ($r=0.44$, $\rho=0.44$), highlighting the effectiveness of induced word embeddings from sub-word (character) information. 
This is not a surprise given that around a third of the rare words in the RW dataset are plural or -\textit{ed} forms which can be easily handled by resorting to the embedding of their singular or uninflected forms. 
For instance, \textit{kindergarteners} and \textit{postponements} are highly similar to their singular forms and the semantics of \textit{encrusted} and \textit{entrapped} can be estimated to a good extent from \textit{encrust} and \textit{entrap} which are relatively more frequent terms. 
None of the other models in the table have access to this information. 
However, as mentioned earlier, the sub-word backoff strategy might not be effective for single-morpheme words and exocentric compounds, which in a real-world scenario account for the most frequent cases of unseen words and can be effectively handled by our model.

\paragraph{Reliability of the RW dataset.}

The Stanford Rare Word Similarity dataset has been regarded as a standard evaluation benchmark for rare word representation and similarity, and as such it is included in the experiments of this paper.
However, the variance across the scores provided by different annotators for the same pair is generally high in this dataset.
This is mainly due to the reliance of the dataset on crowdsourcing without having rigorous checkpoint on the raters.
As also highlighted by \newcite{card660}, the low-confidence annotations are also reflected by contradictory instances, such as the two (almost) identical pairs \textit{tricolour}-\textit{flag} and \textit{tricolor}-\textit{flag} which have received the two very different scores of 5.80 and 0.71.
Hence, further improvements on the dataset (over the \textsc{w2v-gn} baseline), provided by different techniques, cannot be meaningfully interpreted.
Given the unreliability of the benchmark, in the following section, we provide an alternative evaluation based on standard (common) word similarity benchmarks.


\begin{table*}[t!]
\setlength{\tabcolsep}{9pt}
\begin{center}
\scalebox{1}
{
\begin{tabular}{ll cc cc cc cc cc}
\toprule
\multirow{2}{*}{\bf Embedding} & 
\multirow{2}{*}{\bf Setting} &
\multicolumn{2}{c}{\bf RG-65} &
\multicolumn{2}{c}{\bf SimLex-999} &
\multicolumn{2}{c}{\bf MEN-3000} &
\multicolumn{2}{c}{\bf SimVerb-3500} &
\multicolumn{2}{c}{\bf WS-353 Sim} \\

\cmidrule(lr){3-4}
\cmidrule(lr){5-6}
\cmidrule(lr){7-8}
\cmidrule(lr){9-10}
\cmidrule(lr){11-12}

& &
$r$ &  $\rho$   &
$r$ &  $\rho$   &
$r$ &  $\rho$   &
$r$ &  $\rho$   &
$r$ &  $\rho$   \\

\midrule












\multirow{4}{*}{Initial word2vec}

&\multirow{1}{*}{$T=10$} 
&0.40 &0.42 &0.15 &0.12 &0.46   &0.45   &0.07   &0.08   &0.53   &0.54 \\ 

&\multirow{1}{*}{$T=20$} 
&0.54 &0.56 &0.22 &0.21 &0.53   &0.52   &0.12   &0.11   &0.63   &0.62 \\ 

&\multirow{1}{*}{$T=50$} 
&0.63 &0.63 &0.26 &0.24 &0.63   &0.62   &0.15   &0.15   &0.68   &0.69 \\ 

&\multirow{1}{*}{$T=100$} 
&0.68 &0.69 &0.30 &0.28 &0.65   &0.64   &0.19   &0.18   &0.73   &0.73 \\ 

\cmidrule(lr){1-12}


\textsc{Align} & $T=0$ 
&\bf 0.86 &\bf 0.88 &\bf 0.40 &\bf 0.37 & 0.65   & 0.66   &\bf 0.42   &\bf 0.39   & 0.71   & 0.69  \\ 

LSTM & $T=0$
&0.52 &0.57 &0.19 &0.19 &0.19 &0.20 &0.28 &0.29 &0.18 &0.21 \\

Additive & $T=0$ & 0.56 & 0.59 &0.17 & 0.13 & 0.24& 0.23 &0.21 &0.20 &0.31 &0.32\\

SemLand & $T=0$ & 0.52 & 0.53 & 0.22 & 0.20 & 0.38 & 0.38 & 0.23 & 0.22 & 0.43 & 0.40 \\

\midrule
FastText & $T=0$
&0.77 &0.80 &0.32 &0.32 &\bf 0.76 &\bf 0.76 &0.22 &0.21 &\bf 0.74 &\bf 0.73 \\


\bottomrule

\end{tabular}
}
\end{center}
\caption{\label{table:sim-sim}
Results of corpus-based and enhanced embeddings in the simulated rare word similarity setting.}
\end{table*}

\subsection{Simulated Rare Word Similarity}
\label{sec:sim_exp}

For a word similarity dataset to be suitable for this evaluation, it has to contain words that are infrequent in generic texts.
However, most of the existing standard word similarity datasets contain only high frequency words, which makes them unsuitable for evaluating rare word representation techniques.
To work around this limitation, we follow \citep{SergienyaSchutze:2015} and leverage corpus downsampling in order to artificially transform standard word similarity datasets to rare word similarity benchmarks.
This enables us to evaluate our embedding induction technique on a variety of standard datasets. 

\paragraph{Experimental setup.} Let $T$ be the \textit{rarity threshold}, i.e., the expected occurrence frequency of an artificial rare word in the training text corpus.
We process the original text corpus in order to guarantee that each word in the similarity dataset appears at most $T$ times in the training corpus.
This can be achieved by replacing all but $T$ occurrences of the word with another unique token (e.g., the word concatenated by some unique character).
As a result of this procedure, we obtain a corpus for each $T$ value and for each dataset.
Training word embeddings on these corpora simulates a setting in which all the words in the word similarity dataset are rare as they occur infrequently in the training corpus.
Except from the corpus downsampling step, the experimental setup is similar to that of the previous experiment.

\paragraph{Datasets.} For this experiment, we opted for five standard word similarity datasets:
RG-65 \cite{RG65:1965}, SimLex-999 \cite{Hilletal:2015}, MEN \cite{Brunietal:2014}, WordSim-353 similarity subset \cite{Agirreetal:2009}, and SimVerb-3500 \cite{Gerzetal:2016} which contains verbs only.

\paragraph{Results.} 
Table \ref{table:sim-sim} lists correlation performance results on the five datasets and for four different values of $T$ (10, 20, 50, and 100) for the initial downsampled \textsc{w2v-wp} embeddings\footnote{Obviously, for $T=0$, word2vec would be unable to learn any embeddings, hence we do not show that setting.} as well as for enhanced embeddings using different techniques for $T=0$ (unseen word setting).
As expected, there is a steady improvement for the corpus-based embeddings with increasing values of $T$.
On all the datasets and according to both evaluation measures, \textsc{Align} significantly improves over the three other WordNet-based approaches.
Interestingly, our induced embeddings consistently outperform corpus embeddings which are constructed with $T=10$, $20$, and $50$ on all the datasets and are often better or on par with $T=100$.
This means that our approach can produce embeddings that are as reliable as those corpus embeddings that are computed based on 100 occurrences.
This is important as around 80\% of the words in the vocabulary of the Wikipedia corpus appear fewer than 50 times in the whole corpus (see Footnote \ref{foot:vocab}).
Moreover, surprisingly, on the SimVerb dataset the induced embeddings perform significantly better than the corpus-based embeddings, even at $T=100$.
This shows the superior quality of the induced verb embeddings, thanks to the hand-crafted part-of-speech-specific knowledge encoded for them in WordNet.

Similarly to the previous experiment, FastText proves to be a competitive baseline, outperforming our induced embeddings on two datasets.
However, again, we note that FastText benefits from the advantage of having access to all
plural forms of these (originally frequent) downsampled words in the training dataset, which might not establish a fair comparison.
The simulated rare word similarity datasets address the unreliability issue of Stanford RW but still do not represent a real-world rare word scenario.
Ideally, such a dataset would contain named entities, domain-specific terms or other uncommon words that tend to appear infrequently in generic text corpora (which are often used for training word embeddings).
We believe that rare word representation research requires such a high quality benchmark for more rigorous evaluations.
We leave the possibility of the creation of such datasets to future work.

\subsection{Evaluation in Downstream Tasks}
\label{sec:downstream}

We were also interested in having an \textit{in-vivo} evaluation of the reliability of our induced embeddings in a real-world NLP system.
Given that currently the most important application of word embeddings is in the initialization of the input layer in neural networks, we opted for a standard neural system as our evaluation benchmark.

\begin{table*}[t!]
\setlength{\tabcolsep}{11pt}
\begin{center}
\scalebox{1}
{
\begin{tabular}{ll ccccc ccc}
\toprule
\multirow{2}{*}{\bf Initialization} & 
\multirow{2}{*}{\bf Setting} &
\multicolumn{5}{c}{\bf Sentiment Analysis} &
\multicolumn{3}{c}{\bf Topic Categorization} \\

\cmidrule(lr){3-7}
\cmidrule(lr){8-10}

& &
PL04 &  PL05   &
RTC &   IMDB   &
Stanford &  BBC   &
NG &  OH   \\

\midrule

\multirow{2}{*}{X $=0\%$}

& Initial
&\bf 66.2  &75.4  & \bf79.7  &85.4  &80.4    & \bf 96.7    &86.5	 & 27.8 \\

 & +\textsc{Align}
&63.7  & \bf75.6 &79.4   & \bf86.8 & \bf80.5   & 96.5     &\bf87.0	 &\bf29.3  \\

\cmidrule(lr){1-10}

\multirow{2}{*}{X~$=$~20\%}

& Initial
& \bf59.1  & 67.2  &63.8   &71.1   & 70.1   & 93.1     &67.4	 & 16.4  \\

 & +\textsc{Align} 
& 58.9   & \bf69.9   &\bf 74.5  &\bf79.3   & \bf 77.6  &\bf 95.1  &\bf 80.3	 &\bf 25.7  \\

\cmidrule(lr){1-10}

\multirow{2}{*}{X $=40\%$}

& Initial
&\bf 56.2   &63.5   &62.7   & 70.3  & 66.1   & 91.0     &62.8	 & 15.7  \\

 & +\textsc{Align} 
&55.6   & \bf 68.0  & \bf 74.5  & \bf81.8  &\bf 76.2   &\bf 94.5  &\bf 79.7	 &\bf 28.5  \\













\bottomrule

\end{tabular}
}
\end{center}
\caption{\label{table:downstream}
Accuracy performance on eight datasets for sentiment analysis and topic categorization. The best results for each setting are shown in bold. NG and OH stand for Newsgroups and Ohsumed, respectively.}
\end{table*}

\paragraph{Experimental setup.} We experimented with a neural text classification system applied to two tasks: sentiment analysis (binary classification) and topic categorization (multi-class classification).
The embedding layer of this system is initialized with pre-trained word2vec embeddings. Let $L$ be the vocabulary of a given dataset.
We dropped the pre-trained corpus embeddings 
for $X\%$ of the words in $L$ and replaced them with our induced embeddings.
We experimented with three $X$ values: 0 (in which we used all the corpus embeddings to initialize the layer; new embeddings were induced to further improve coverage for those words missing in corpus embeddings' vocabulary), 20 and 40 (in which, respectively, 20\% and 40\% of corpus embeddings were dropped, i.e., their corresponding words were treated as out of vocabulary).
We were mainly interested in observing if the induced embeddings, first, could improve over corpus embeddings and, second, were able to re-gain system performance lost when dropping a part of the corpus embeddings.
In all settings the embedding layer was not updated during training (static).
This allows us to have a direct evaluation on the reliability of embeddings, independently from any updates and alteration they can undergo during training.
In each configuration we repeat the experiment three times and report the average performance.

\paragraph{Text classification system.} In our experiments, we used a CNN text classifier which is similar to that of \newcite{kim2014convolutional}.
The only difference is that in our model, instead of directly inputting the pooled features from the convolutional layer to a fully connected softmax layer, they are first passed through a recurrent layer in order to enable a better capturing of long-distance dependencies.
Specifically, as our recurrent layer we used LSTM \cite{hochreiter1997long}.

\paragraph{Datasets.} For sentiment analysis we used five standard datasets, including PL04 \cite{Pang+Lee:04a}, PL05 \cite{Pang+Lee:05a},\footnote{Both PL04 and PL05 are obtained from \url{http://www.cs.cornell.edu/people/pabo/movie-review-data/}}
RTC\footnote{\url{http://www.rottentomatoes.com}}, and IMDB \cite{maas-EtAl:2011:ACL-HLT2011} which are all binary datasets (with positive and negative labels) containing snippets of or full movie reviews.
We also experimented with Stanford Sentiment dataset \cite{SocherEtAl2013:RNTN} which associates phrases with values that denotes their sentiments.
To be consistent with the other four datasets' binary classification setting, we removed the neutral phrases with scores 0.4 to 0.6 and considered the reviews with values below 0.4 as negative and above 0.6 as positive.
For the topic categorization task we used two newswire datasets:
The BBC news dataset
CR \footnote{\url{http://mlg.ucd.ie/datasets/bbc.html}} 
\cite{greene2006practical} and  Newsgroups \cite{lang1995newsweeder} with 5 and 20 classes, respectively.
We also experimented with a domain-specific categorization dataset: Ohsumed\footnote{\url{ftp://medir.ohsu.edu/pub/ohsumed}}, which contains medical texts categorized into 23 classes. 
\paragraph{Results.} Table \ref{table:downstream} shows the results.
We report classification accuracy for the baseline system (``Initial'') which is initialized by full ($X=0\%$) or partial ($X>0\%$) corpus-based embeddings, and for the enhanced systems with additional induced embeddings (``+\textsc{Align}'').
Generally, the enhancement proves to be beneficial as it provides improvements in most of the configurations across the eight datasets.
In the $X=0\%$ setting, the improvement is particularly noticeable for the IMDB, Newsgroup and Ohsumed datasets which have a fair portion of their vocabularies not covered by word2vec embeddings. 
However, lower or no improvement is observed for other datasets (particularly, PL04) whose vocabularies are largely covered by the corpus embeddings.
In the $X>0\%$ settings, the performance of the baseline system drops significantly on most datasets.
In the 20\% setting, which is the closest to a real-world scenario, the enhanced system can recover a large part of the lost performance on most of the datasets.
The same trend is observed for $X=40\%$.
Interestingly, on the Ohsumed dataset, which belongs to the medical domain, the enhanced system gets close to the initial system initialized by corpus embeddings.
This is a strong indication of the effectiveness of our approach in filling lexical gaps for specific domains.
Overall, the results show that our induced embeddings, though not sufficient to replace corpus embeddings for frequent words, can significantly improve over infrequent or unkown embeddings, particularly for specific domains.

\section{Related Work}
\label{sec:related_work}

Given its importance, unseen word representation has attracted considerable research attention for the past few years.
Earlier techniques have mainly focused on improving distributional models for better handling of infrequent words \cite{SergienyaSchutze:2015}, or on inducing embeddings for morphological variations \cite{alexandrescu-kirchhoff:2006,luong-socher-manning:2013,lazaridouetal:2013,Botha2014,soricut-och:2015}.
%
The latter branch often utilizes a morphological segmenter, such as Morfessor \cite{CreutzLagus:2007}, in order to break inflected words into their components and to compute representations by extending the semantics of an unseen word's morphological variations.

More recently, character-based models have garnered a lot of attention.
In these models words are broken down into subword units and characters \cite{fasttext-subword:TACL999}, usually irrespective of their morphological structure.
An unseen word's representation is induced by combining the information for its subword units; for instance, by averaging the
vector representations of its constituent character n-grams as done by FastText \cite{fasttext-subword:TACL999}.
Character-based models have been successfully tested in different NLP tasks, including language modeling \cite{Sutskever:2011,Graves:2013}, part-of-speech tagging \cite{DosSantosetal:2014,Lingetal:2015} and syntactic parsing \cite{Ballesterosetal:2015}.
However, all these techniques fall short of inducing representations for single-morpheme words that are not seen frequently during training as they base their modeling on information available from sub-word units. In contrast, our alignment-based model can also induce embeddings for single-morpheme words that are infrequent or unseen in the training data, such as domain-specific entities.

Most related to our work are the WordNet-based approaches of \newcite{pilehvar-collier:2017:EACLshort} and \newcite{embeddings-on-fly:2017}. 
The former computes an unseen word's embedding by extracting the set of its semantically similar words (``semantic landmarks'') from WordNet and combining their embeddings, whereas the latter 
trains a recurrent neural network, specifically, an LSTM, to estimate a word's embedding given its definition from WordNet.
Moreover, the additive model of \newcite{Angeliki2017-LAZMWM} is analoguous to the LSTM model (though less complex) and computes an embedding as the centroid of the embedding of the words in its definition.
Despite addressing the single-morpheme word representation limitation of morphological models, these approaches ignore the information encoded in WordNet's lexical-semantic relations.
We improve over these by proposing a model that effectively leverages the semantic network of WordNet.
Our experimental results show that, for the task of embedding induction, structural information can result in a more consistent performance than glosses or similar words.

\section{Conclusions and Future Work}

We presented a methodology for marrying distributional semantic spaces with lexical knowledge bases and applied it to the task of extending the vocabulary of the former with the help of information extracted from the latter.
We showed the reliability of our approach by evaluating the induced embeddings on multiple word similarity benchmarks as well as on a downstream NLP evaluation framework.
In future work, we plan to experiment with domain-specific lexical resources, such as medical ontologies, and study the efficacy of our methodology on adapting downstream NLP systems to new domains.

\section*{Acknowledgments}

We would like to thank the anonymous reviewers for their comments. This research was supported by an EPSRC Experienced Researcher Fellowship (N. Collier, D. Kartsaklis: EP/M005089/1) and an MRC grant (M.T. Pilehvar: MR/M025160/1). We gratefully acknowledge the donation of a GPU from the NVIDIA Grant Program.

\bibliography{acl2018}
\bibliographystyle{aaai}

\appendix

\end{document}